\documentclass[runningheads]{llncs}
\usepackage[T1]{fontenc}
\usepackage{graphicx,verbatim}
\usepackage{amsmath}
\usepackage{booktabs}
\usepackage{multirow}
\usepackage[table]{xcolor} 
\usepackage{diagbox}
\usepackage{amssymb}
\usepackage{bbold}
\usepackage{bbding}
\usepackage{caption}

\usepackage{booktabs}

\usepackage{marvosym}
\usepackage{hyperref}
\usepackage{overpic}
\usepackage{bm}
\usepackage{amsmath,array,arydshln}
\usepackage{multicol}
\usepackage{multirow}
\usepackage{hyperref}
\usepackage{amsfonts}
\usepackage{cite}
\usepackage{ifsym}
\usepackage{subfigure}

\begin{document}
\title{CHILD: Human-in-the-Loop OOD Detection for Safe Clinical Deployment}

%
\author{Jinlun Ye\inst{1,2,4} \and
Kaiyue Lu\inst{1,2,4} \and
Runhe Lai\inst{1,2,4} \and
Xinhua Lu\inst{1,2,4} \and
Jia-Xin Zhuang\inst{3} \textsuperscript{(\Letter)} \and
Ruixuan Wang\inst{1,2,4} \textsuperscript{(\Letter)}
}

\authorrunning{J. Ye et al.}
%
\institute{School of Computer Science and Engineering, Sun Yat-sen Univerisity, Guangzhou,China\and
Peng Cheng Laboratory, Shenzhen, China\and
Hong Kong University of Science and Technology, Hong Kong, China \and
Key Laboratory of Machine Intelligence and Advanced Computing, MOE, Guangzhou, China \\
\email{jzhuangad@connect.ust.hk, wangruix5@mail.sysu.edu.cn}
}



  
\maketitle              
\begin{abstract}
Out-of-distribution (OOD) detection is critical for safe deployment of medical AI systems.
Recently, test-time adaptation (TTA) has emerged as a new paradigm for OOD detection, automatically adjusting detector behavior during deployment.
However, such automatic adaptation mechanisms may raise safety concerns in safety-critical clinical environments.
While physician oversight can mitigate these risks, it is resource-intensive and must be judiciously allocated.
To reconcile safety with efficiency,
we propose CHILD, a training-free framework designed to enhance streaming OOD detection via sparse human feedback.
Operating under strict budget constraints, 
CHILD employs an adaptive risk-aware sample selection mechanism to pinpoint only the most decision-uncertain samples for review. Crucially, it maximizes the utility of this sparse feedback through a retrieval-based score calibration module, which refines model predictions using a compact feature cache without any parameter updates.
Extensive experiments on four medical benchmarks demonstrate that CHILD turns limited supervision into significant reliability gains: with a sparse feedback budget of only 5\%, it reduces the average FPR95 from 72.63\% to 60.26\% and improves AUROC from 75.53\% to 81.85\%, consistently outperforming state-of-the-art baselines.
Our code is publicly available at \url{https://github.com/figec/CHILD}.


\end{abstract}

\section{Introduction}
Deep learning models demonstrate strong performance in medical diagnosis when test samples belong to the known disease categories seen during training (in-distribution, ID). However, real-world clinical deployment inevitably exposes such systems to unseen disease categories, namely out-of-distribution (OOD) samples, where models may produce overconfident yet incorrect predictions~\cite{DBLP:conf/cvpr/NguyenYC15,msp}, posing significant clinical risks. Consequently, OOD detection is critical for the safe deployment of medical artificial intelligence.

Existing research on OOD detection can be broadly categorized into two conventional paradigms. One line of work focuses on post-hoc scoring~\cite{msp,DBLP:conf/icml/HendrycksBMZKMS22, energy, react, ash,logitGap,cadref}, where OOD scores are derived from model outputs or intermediate feature representations without modifying the training procedure.
Another line explores training-time OOD-aware learning~\cite{locoop,ospcoop,fa,tagfog,fodfom,multi}, aiming to improve model reliability by regularization strategies or synthetic OOD exposure during training. 
More recently, TTA~\cite{rtl,oodd,modelfree, noisy,dcac, ttl} has emerged as a new paradigm. By leveraging unlabeled test data observed during deployment, TTA methods adjust model behavior at inference time and have been shown to achieve better OOD detection performance compared to static methods.

However, directly applying TTA methods to clinical deployment raises safety concerns. TTA methods adapt detection behavior automatically based solely on unlabeled test data, leading to unpredictable decision adjustments and progressive error accumulation~\cite{histpt}.
At the same time, physician review of uncertain or high-risk cases is not an additional burden but an integral component of routine workflows.
Such naturally available and reliable clinical supervision motivates a safer strategy: instead of relying entirely on automatic adaptation, selectively incorporating limited physician feedback can guide decision adjustment at inference time in a controlled and clinically aligned manner.
Recent work~\cite{humun_out} has explored the use of human feedback for OOD detection through additional training, but assumes offline access to the entire test set for feedback allocation and decision adjustment. In real clinical deployment, data arrive sequentially in a streaming manner, and decisions must be made online using only past observations, rendering offline methods impractical.

Therefore, we propose \textbf{C}linical \textbf{H}uman-\textbf{I}n-the-\textbf{L}oop \textbf{D}ecision (\textbf{CHILD}), a training-free human-in-the-loop OOD detection framework motivated by clinical deployment.
CHILD consists of two components: (i) \emph{Risk-Aware Sample Selection}, 
which adaptively identifies decision-uncertain regions in the score space and allocates limited physician feedback to samples with the highest potential clinical risk; and
(ii) \emph{Score Calibration}, which 
maintains compact feature caches of 
physician-reviewed ID/OOD samples
and adjusts OOD scores via similarity-driven calibration, without any parameter updates.
Our contributions are summarized as follows:
\begin{enumerate}
    \item 
    We formulate a streaming human-in-the-loop OOD detection setting with strict feedback budget constraints, capturing the sequential decision-making requirements relevant to clinical deployment.
    \item 
    We propose CHILD, a training-free and plug-and-play framework that selectively queries physician feedback for high-risk samples and incorporates it through lightweight score-level calibration.
    \item 
    Extensive experiments on four medical benchmarks demonstrate that CHILD consistently improves performance with minimal feedback, reducing FPR95 by over 12\% on average when combined with a state-of-the-art detector.
\end{enumerate}

\section{Method}

\subsection{Preliminaries and Problem Setup}

\noindent \textbf{OOD Detection.}
Given a test sample $\mathbf{x}_t$, an OOD detector $f(\cdot)$ produces a scalar score $ s_t = f(\mathbf{x}_t)$.
OOD detection is typically performed by comparing the score $s_t$ with a decision threshold $\lambda$, i.e.,
\begin{equation}
\label{eq:ood}
G(\mathbf{x}_t) =
\begin{cases}
\text{ID}, & s_t \ge \lambda, \\
\text{OOD}, & s_t < \lambda .
\end{cases}
\end{equation}

\noindent \textbf{Human-in-the-Loop Streaming OOD Detection Setting.}
In this work, we consider a deployment setting where test samples arrive sequentially as a stream $\{\mathbf{x}_t\}_{t=1}^{T}$, with ID and OOD samples mixed and unlabeled.
During deployment, the system is allowed to request a limited amount of human feedback under a strict budget constraint.
Specifically, at most $\rho\%$ of test samples can be reviewed by physicians.
Rather than requiring physicians to directly determine the model-defined ID/OOD classes, they provide a diagnostic category or an intended-scope judgment, which is then mapped to ID/OOD according to the predefined set of ID classes associated with the model's intended use.
In our experiments, benchmark annotations serve as a controlled proxy for such physician-confirmed feedback.
We use $q_t \in \{0,1\}$ to indicate whether physician feedback is requested for sample $x_t$, and enforce the budget constraint
\begin{equation}
\label{eq:budget}
\sum_{t=1}^{T} q_t \le B,
\end{equation}
where $B = \lfloor \rho T \rfloor$ denotes the total feedback budget.
Human feedback is acquired incrementally as samples arrive, 
and the system has no access to the complete test set in advance.

\subsection{Overview}
\begin{figure}[t]
\begin{center}
\includegraphics[width=0.9\textwidth]{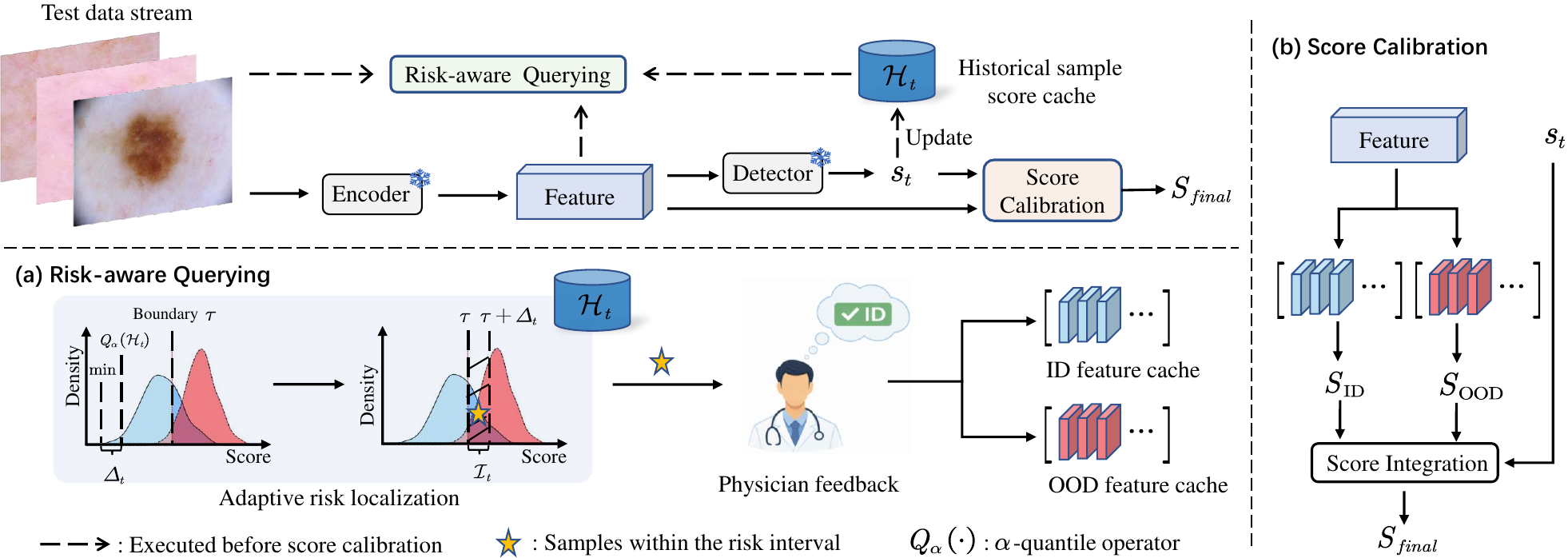}
\end{center}
   \caption{
Overview of the proposed framework. 
During deployment, samples falling within the adaptive risk interval are queried for feedback. 
The collected feedback is stored and subsequently used to calibrate the OOD score of incoming samples.
}
\label{figure_framework}
\end{figure}

Under the streaming OOD detection setting, we propose \textbf{C}linical \textbf{H}uman-\textbf{I}n-the-\textbf{L}oop \textbf{D}ecision (\textbf{CHILD}), a training-free human-in-the-loop framework designed for clinical deployment.
As illustrated in Figure~\ref{figure_framework}, CHILD 
operates on top of an arbitrary base OOD detector and 
consists of two interleaved components that operate jointly as each sample arrives.
The first component, \emph{Risk-Aware Sample Selection}, allocates limited physician feedback to decision-uncertain samples that are associated with higher potential clinical risk.
The second component, \emph{Score Calibration}, incorporates the obtained ID/OOD feedback as decision-time priors to calibrate the outputs of the base OOD detector during inference.

\subsection{Risk-Aware Sample Selection}
Under a limited feedback budget, physician review should be allocated to samples that are most likely to reduce high-risk errors in a streaming environment. 
To this end, the proposed module maintains an online estimate of the score distribution and localizes high-risk regions for targeted feedback allocation (detailed below).

\textbf{Online Score Accumulation.}
During deployment, the system continuously records OOD scores 
produced by the base detector as test samples arrive.
Let $\mathcal{H}_t = \{s_i\}_{i=1}^{t}$ denote the set of accumulated historical scores up to time step $t$. 
These scores provide an evolving empirical estimate of the score distribution under the current deployment environment and form the basis for identifying decision-uncertain and potentially high-risk regions.

\textbf{Adaptive Risk Sample Localization.}
In medical OOD detection, the most critical failure mode arises when OOD samples are misclassified as ID, a situation that typically occurs near regions of decision uncertainty where the base detector struggles to separate ID from OOD samples. Compared with confirming high-confidence samples far from the decision boundary, allocating limited human feedback to such decision-uncertain samples is more likely to influence system decisions and achieve greater risk reduction.

Accordingly, we aim to localize decision-uncertain regions along the one-dimensional OOD score axis by estimating a time-evolving reference boundary $\tau_t$.
 The accumulated OOD scores form an empirical one-dimensional distribution, from which a boundary separating lower- and higher-score regions can be estimated directly in score space.
To this end, the Otsu~\cite{otsu} thresholding method is adopted. 
For a candidate threshold $\tau$, the score distribution is divided into two groups, and the between-class variance $\sigma_b^2(\tau)$ quantifies the separation between their mean scores. 
The threshold that maximizes $\sigma_b^2(\tau)$ is selected as the reference boundary, which typically lies between distinct score regimes, making it suitable for approximating the current ID/OOD decision boundary: 
\begin{equation}
\label{eq:boundary}
\tau_t = \arg\max_{\tau}\ \sigma_b^2(\tau) \,.
\end{equation}
The boundary $\tau_t$ is continuously updated as streaming data arrive, allowing the estimate to adapt to distributional changes during deployment.
To operationalize feedback querying, $\tau_t$ is used as a reference point and define a risk query interval around it. 
To ensure that the width of the risk interval adapts to the evolving score distribution, the width term $\Delta_t$ is defined as a function of $\mathcal{H}_t$, i.e.,
\begin{equation}
\label{eq:scope}
\Delta_t = Q_{\alpha}(\mathcal{H}_t) - \min(\mathcal{H}_t) \,,
\end{equation}
where $Q_{\alpha}(\cdot)$ denotes the $\alpha$-quantile operator, and $\alpha \in (0,1)$ is typically chosen to be close to $0$.
 This formulation defines the interval width with respect to the lower extreme of the score distribution while using a quantile-based statistic to provide a robust and adaptive characterization of the distributional spread over time.
 In medical OOD deployment, risk is asymmetric in score-based decisions, as the most critical failure occurs when OOD samples are incorrectly accepted as ID.
We then construct an asymmetric risk interval $\mathcal{I}_t$, i.e.,
\begin{equation}
\label{eq:interval}
\mathcal{I}_t = [\,\tau_t,\; \tau_t + \Delta_t\,] \,,
\end{equation}
which expands coverage toward the more ID-like side of the boundary compared with a symmetric interval.
This design increases the likelihood of capturing samples whose misclassification would lead to more severe clinical consequences under the same feedback budget.

\textbf{Physician Feedback and Sample Caching.}
For samples within the risk interval, the system requests physician feedback when the budget permits.
The provided diagnostic category or intended-scope judgment is mapped to an ID/OOD label according to the predefined ID class set.
Each feedback instance is treated as high-confidence prior information. 
The feature $\phi(\mathbf{x})$ of a sample mapped to ID or OOD is stored in $\mathcal{C}_{\text{ID}}$ or $\mathcal{C}_{\text{OOD}}$, respectively, for subsequent score calibration.
Their total size satisfies
$|\mathcal{C}_{\text{ID}}|+|\mathcal{C}_{\text{OOD}}|\leq B$.
Throughout this process, no model parameters are updated.

\subsection{Score Calibration}

Given the feedback caches $\mathcal{C}_{\text{ID}}$ and $\mathcal{C}_{\text{OOD}}$, our goal is to refine OOD decisions at inference time without any test-time training. To achieve this, we perform lightweight score-level calibration that directly leverages similarity to physician-confirmed samples.
For a new test sample $\mathbf{x}_t$, the base detector first produces the original OOD score $s_t$. 
The maximum cosine similarity between $\mathbf{x}_t$ and each feedback cache is then computed as follows:
\begin{equation}
S_{\text{ID}}(\mathbf{x}_t) = \max_{\phi(\mathbf{x}) \in \mathcal{C}_{\text{ID}}} \cos(\phi(\mathbf{x}_t), \phi(\mathbf{x})), \quad
S_{\text{OOD}}(\mathbf{x}_t) = \max_{\phi(\mathbf{x}) \in \mathcal{C}_{\text{OOD}}} \cos(\phi(\mathbf{x}_t), \phi(\mathbf{x})),
\end{equation}
Then the similarity difference $\delta_t =S_{\text{ID}}(\mathbf{x}_t) - S_{\text{OOD}}(\mathbf{x}_t)$ can indicate whether the sample is closer to confirmed instances and, therefore, is used as a soft calibration term to adjust the base score. However, in certain cases,
a test sample may be highly similar to a previously confirmed sample, 
indicating strong semantic consistency; in such situations, relying solely on the soft calibration may lead to insufficient calibration strength. 
To provide a more decisive adjustment under clear semantics, we introduce a hard-assignment mechanism:
\begin{equation}
y_t = \mathrm{sign}\!\left(S_{\text{ID}}(\mathbf{x}_t) - S_{\text{OOD}}(\mathbf{x}_t)\right),
\end{equation}
which reflects whether the sample is more similar to confirmed ID or OOD instances. 
To ensure that hard assignment is applied exclusively under reliable evidence, we further introduce a hard-assignment gate:
\begin{equation}
h_t =  \mathbb{1}\!\left[\max\{S_{\text{ID}}(\mathbf{x}_t), S_{\text{OOD}}(\mathbf{x}_t)\} \ge \eta \right],
\end{equation}
where $\mathbb{1}[\cdot]$ denotes the indicator function that takes value 1 if the condition is satisfied and 0 otherwise, and $\eta$ is the semantic confidence threshold.
The resulting hard-assignment term is thus denoted as $H_t = h_t y_t$.
Based on these quantities, the final calibrated score is defined as
\begin{equation}
S_{final}
= s_t + \beta\Big( \delta_t  + \, H_t \Big) \,,
\end{equation}
where $\beta$ serves as a scaling factor that aligns the magnitude of the similarity-based calibration terms with the base detector score, while also controlling the overall calibration strength.
This formulation unifies soft calibration and hard assignment within a single score-level adjustment.
Higher similarity to physician-confirmed OOD samples reduces the calibrated score, pushing the decision toward OOD, whereas higher similarity to physician-confirmed ID samples increases the calibrated score, reinforcing the ID decision.
In this way, limited human feedback is translated into reliable decision refinement.

\section{Experiment}
\subsection{Experiment Setup}
\textbf{Datasets.}
Following prior evaluation protocols~\cite{hvl,are_we, gobal}, we evaluate CHILD on SD-198~\cite{skin198}, ISIC 2019~\cite{isic2019}, NCT-CRC~\cite{nct_crc}, and BreakHis~\cite{breakhis}, covering diverse clinical imaging scenarios.
For SD-198, following standard practice~\cite{skin40},
40 categories are designated as ID categories (referred to as Skin-40), while the remaining 158 categories are treated as OOD data.
For ISIC 2019, NV, MEL, DF, and VASC are selected as ID categories, forming a long-tailed ID setting. This setting is referred to as ISIC-4, while the remaining four categories serve as OOD data.
For NCT-CRC and BreakHis, we follow commonly used OOD detection setups with multiple ID/OOD splits~\cite{are_we}, and the final performance is reported by averaging results across all splits.

\noindent \textbf{Implementation Details.}
Unless otherwise specified, all experiments are conducted using a standard image classifier trained on the ID training set with cross-entropy loss.
During deployment, the human feedback budget is set to $\rho=5\%$. For risk interval construction, the quantile level is fixed to $\alpha=1\%$. In the score calibration module, the calibration coefficient is set to $\beta=0.08$, and the semantic confidence threshold is set to $\eta=0.8$.
For evaluation under the streaming setting, decision calibration is first applied at each time step using feedback accumulated from previous samples, after which the system determines whether the current sample should be queried for human feedback. Any feedback obtained from the current sample is used only for calibrating subsequent samples, ensuring a strictly causal evaluation without information leakage.

\noindent \textbf{Comparison methods.} 
All methods are evaluated using a shared ResNet-50 backbone. OOD detectors are constructed with representative post-hoc scoring functions (MSP~\cite{msp}, Energy~\cite{energy}, React~\cite{react},
LogitGap~\cite{logitGap}, CADRef~\cite{cadref}) and TTA methods (Online RTL~\cite{rtl}, DCAC~\cite{dcac}, OODD~\cite{oodd}). 
Under the same feedback budget, we construct two additional baselines. RSB adopts random querying with physician feedback and applies only the soft calibration term $\delta_t$. RPB replaces physician feedback with pseudo labels~\cite{noisy} while retaining both soft and hard calibration.
Performance is measured by FPR95 and AUROC.

\begin{table}[!t]
\centering
\caption{Performance comparison on the four benchmarks. All values are percentages.
Lower FPR95 and higher AUROC are better. 
Best results are in \textbf{bold} and the second-best results are \underline{underlined}. }
\label{tab:imagenet-1k}
\resizebox{\textwidth}{!}{
\begin{tabular}{lccccccccccc}
\toprule
\multicolumn{1}{c}{ID} 
& \multicolumn{2}{c}{ISIC-4} 
& \multicolumn{2}{c}{Skin-40} 
& \multicolumn{2}{c}{NCT-CRC} 
& \multicolumn{2}{c}{BreakHis} 
& \multicolumn{2}{c}{\textbf{Average}} \\
\cmidrule(lr){1-1} \cmidrule(lr){2-3} \cmidrule(lr){4-5} \cmidrule(lr){6-7} \cmidrule(lr){8-9} \cmidrule(lr){10-11}
\multicolumn{1}{c}{Method}   & FPR95$\downarrow$ & AUROC$\uparrow$ & FPR95$\downarrow$ & AUROC$\uparrow$ & FPR95$\downarrow$ & AUROC$\uparrow$ & FPR95$\downarrow$ & AUROC$\uparrow$ & FPR95$\downarrow$ & AUROC$\uparrow$ \\
\midrule
MSP~\cite{msp}             & 82.21 & 73.38 & 86.64 & 68.96 & 69.59 & 81.14 & 88.80 & 67.76 & 81.81 & 72.81 \\
Energy~\cite{energy}       & 73.96 & 74.76 & 84.15 & 70.70 & 62.32 & 81.98 & 78.77 & 68.18 & 74.79 & 73.90 \\
ReAct~\cite{react}         & \underline{69.82} & \underline{78.70} & 85.85 & 71.74  & 62.27 & 85.91 & 80.97 & 66.31 & 74.72 & 75.66 \\
LogitGap~\cite{logitGap}   & 82.57 & 73.20 & 83.96 & 71.98 & 72.41 & 80.82 & 90.76 & 67.52 & 75.74 & 74.12 \\
CADRef~\cite{cadref}       & 77.14 & 73.17 & 85.76 & 58.11 & 52.72 & 84.70 & 88.28 & 60.47 & 75.97 & 69.11 \\
Online RTL~\cite{rtl}     & 78.70 & 64.68 & 86.71 & 68.81 & 63.78 & 80.43 & 86.96 & 62.01 & 79.03 & 68.98 \\
DCAC~\cite{dcac}          & 81.36 & 73.27 & 78.75 & 75.06 & 70.31 & 80.53 & 82.09 & 68.52 & 78.12 & 74.34 \\
OODD~\cite{oodd}          & 85.37 & 71.36 & 75.72 & 74.87 & 51.31 & 83.73 & \underline{78.13} & 72.19 & 72.63 & 75.53 \\ 
\midrule
OODD+RSB     & 83.98 & 72.70 & 75.10 & \underline{75.33} & \underline{44.16} & \underline{86.22} & 80.77 & \underline{73.04} & 71.00 & \underline{76.82} \\
OODD+RPB     & 81.34 & 72.46 & \underline{74.35} & 74.61 & 45.67 & 85.31 & 80.21 & 72.09 & \underline{70.39} & 76.11 \\
\rowcolor{red!10}
\textbf{OODD$+$Ours}     & \textbf{67.74} & \textbf{82.07} & \textbf{73.73} & \textbf{76.20} & \textbf{22.36} & \textbf{93.99} & \textbf{77.24} & \textbf{75.14} & \textbf{60.26} & \textbf{81.85}\\
\bottomrule
\end{tabular}
}
\end{table}

\subsection{Result Analysis}
\begin{figure*}[!t]
    \begin{minipage}[t]{0.64\textwidth}
        \centering
        \includegraphics[width=\textwidth]{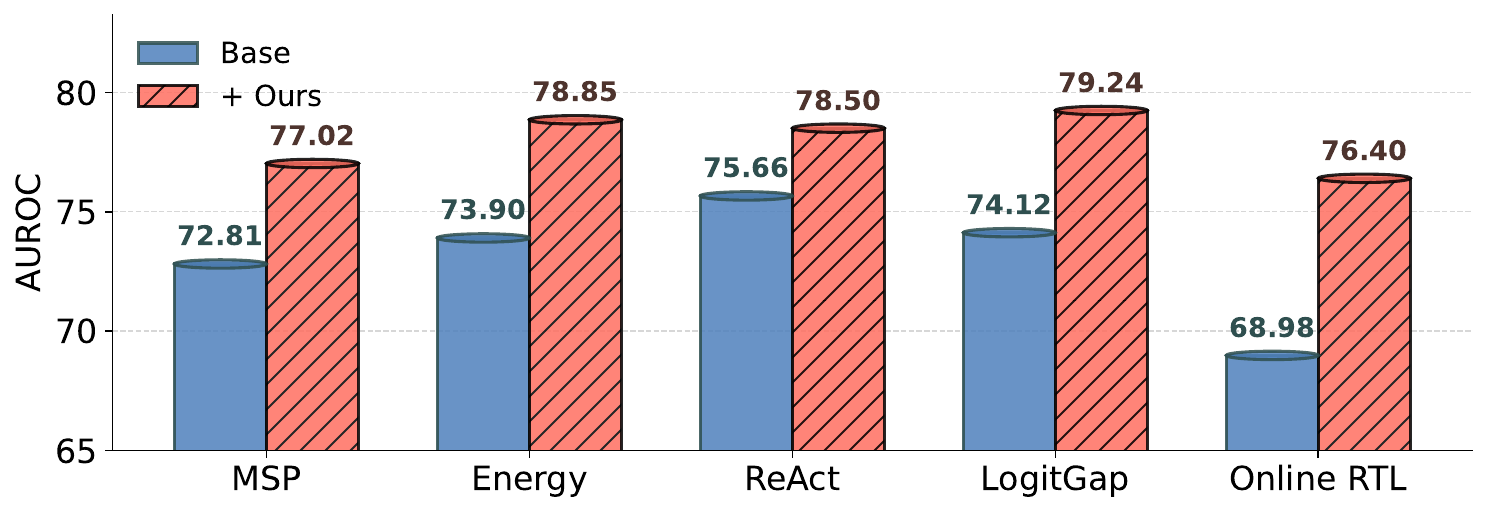}
        \captionof{figure}{Average performance with CHILD integrated into different OOD detectors.}
        \label{fig:overlay_with_other_method}
    \end{minipage}
    \begin{minipage}[t]{0.33\textwidth}
        \centering
        \includegraphics[width=1.0\textwidth]{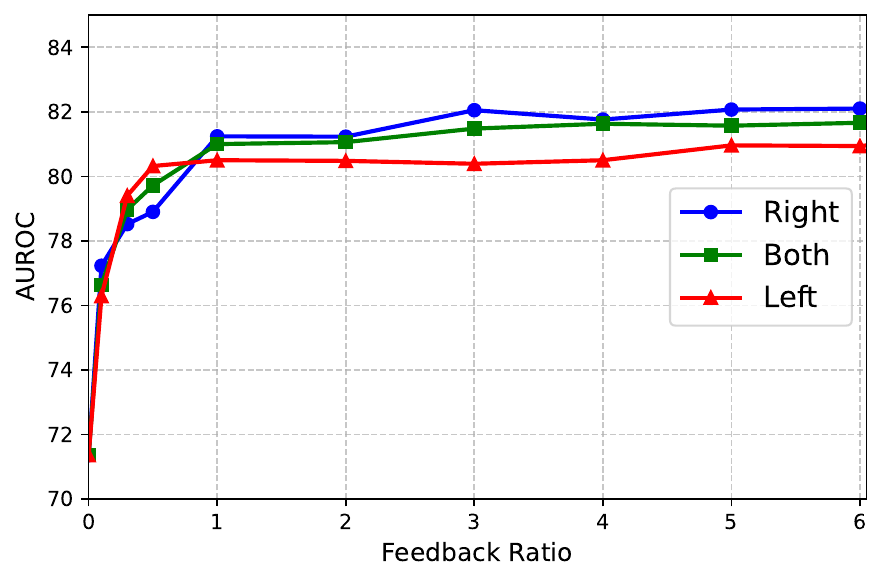}
        \captionof{figure}{Performance when varying feedback ratio.}
        \label{fig:feedback}
    \end{minipage}
\end{figure*}

\noindent \textbf{Efficacy Evaluation.}  
Table~\ref{tab:imagenet-1k} presents the main results on four medical benchmarks.
CHILD achieves the best AUROC and FPR95 when combined with OODD.
Compared with RSB, the improvement verifies that the proposed framework can effectively exploit physician feedback.
Compared with RPB, the gap indicates that pseudo labels provide less reliable guidance than controlled physician supervision in this streaming setting.
Results with other OOD detectors (Fig.~\ref{fig:overlay_with_other_method}) further demonstrate the general applicability of CHILD.

\noindent \textbf{Ablation Study.}
Table~\ref{tab:ablation_ablation} evaluates the impact of Soft Calibration $\delta_t$, Adaptive Risk Sample Localization $\mathcal{I}_t$, and Hard Assignment $ H_t$.
Each component contributes consistent performance gains, and the best results are achieved when all components are enabled, demonstrating their complementary effects.
On NCT-CRC, the gain from $\mathcal{I}_t$ is less pronounced, as samples selected within the risk interval occasionally exhibit similar characteristics, limiting the benefit of boundary-focused querying.

\begin{table}[!t]
        \centering
        \caption{Ablation study of CHILD. In the $\mathcal{I}_t$ column, $\checkmark$ indicates risk-aware sample selection, while \XSolidBrush denotes random selection.} 
        \label{tab:ablation_ablation}
        \resizebox{\textwidth}{!}{
        \begin{tabular}{c p{12pt} c p{11pt} c c p{4pt} | cccc cc cc cc}
        \toprule
        \multirow{2}{*}{\large $\delta_t$} &  & \multirow{2}{*}{\large $\mathcal{I}_t$} & & \multicolumn{2}{c}{$ H_t$}  &
& \multicolumn{2}{c}{ISIC-4} 
& \multicolumn{2}{c}{Skin-40} &  \multicolumn{2}{c}{NCT-CRC} 
& \multicolumn{2}{c}{BreakHis} & \multicolumn{2}{c}{\textbf{Average}}  \\
        \cmidrule(lr){16-17} \cmidrule(lr){8-9} \cmidrule(lr){10-11} \cmidrule(lr){12-13} \cmidrule(lr){14-15} \cmidrule(lr){5-6}
        & & & & $h_t$ & $y_t$ & &  FPR95$\downarrow$ & AUROC$\uparrow$ & FPR95$\downarrow$ & AUROC$\uparrow$ & FPR95$\downarrow$ & AUROC$\uparrow$ & FPR95$\downarrow$ & AUROC$\uparrow$ & FPR95$\downarrow$ & AUROC$\uparrow$  \\
        \midrule
        \rowcolor{gray!20}
        \XSolidBrush & & \XSolidBrush & & \XSolidBrush & \XSolidBrush & & 85.37 & 71.36 & 75.72 & 74.87 & 51.31 & 83.73 & 78.13 & 72.19 & 72.63 & 75.53   \\
        $\checkmark$ & & \XSolidBrush    & & \XSolidBrush & \XSolidBrush  &   & 83.98 & 72.70 & 75.10 & 75.33 & 44.16  & 86.22 & 80.77 & 73.04 & 71.00 & 76.82 \\
        $\checkmark$ & & $\checkmark$   & & \XSolidBrush & \XSolidBrush  &   & 83.26 & 72.87 & 74.28 & 75.36  & 28.23 & 93.34 & 78.38 & 73.54  & 66.03 & 78.77\\
        $\checkmark$ & &  \XSolidBrush  &  &    $\checkmark$ & $\checkmark$ &   & 70.02 & 81.32 & 82.03 & 75.30 & \textbf{17.89} & \textbf{94.91}  & 80.14 & 74.02 & 62.52 & 81.39\\
        $\checkmark$ & & $\checkmark$     &  &  \XSolidBrush & $\checkmark$  & & 70.56 & 79.47 & 73.02 & 75.98 & 28.75 & 91.54 & 77.72 & 74.23 & 62.51 & 80.30\\
        $\checkmark$ & & $\checkmark$     &  &  $\checkmark$ & \XSolidBrush  & & 83.43 & 71.75 & 75.45 & 74.49 & 43.43 & 86.25 & 78.91 & 73.05 & 70.30 & 76.38\\
        $\checkmark$ & & $\checkmark$     &  &  $\checkmark$ & $\checkmark$  & & \textbf{67.74} & \textbf{82.07} & \textbf{73.73} & \textbf{76.20} & 22.36 & 93.99 & \textbf{77.24} & \textbf{75.14} & \textbf{60.26} & \textbf{81.85}\\
        \bottomrule
        \end{tabular}
        }
\end{table}

\noindent \textbf{Budget--Performance Trade-off.}
Fig.~\ref{fig:feedback} illustrates the effect of the feedback budget on ISIC-4 under three interval selection strategies defined relative to the reference boundary $\tau_t$: selecting samples from the left side of $\tau_t$, the right side of $\tau_t$, or symmetrically from both sides.
Overall, right-side selection yields stronger and more stable performance across feedback budgets, as refining the ID-side decision region more effectively corrects high-risk errors and improves ID/OOD score discrimination.
Across all strategies, performance consistently improves as the feedback ratio increases, with AUROC approaching saturation at around $1\%$ feedback.
Beyond $5\%$ feedback, further budget increases bring only marginal gains, as additional samples within the quantile-defined risk interval tend to provide increasingly redundant calibration signals.
These results indicate that the proposed framework is effective and robust under limited feedback budgets.

\begin{figure*}[!t] 
        \begin{center}
        \includegraphics[width=0.85\textwidth]{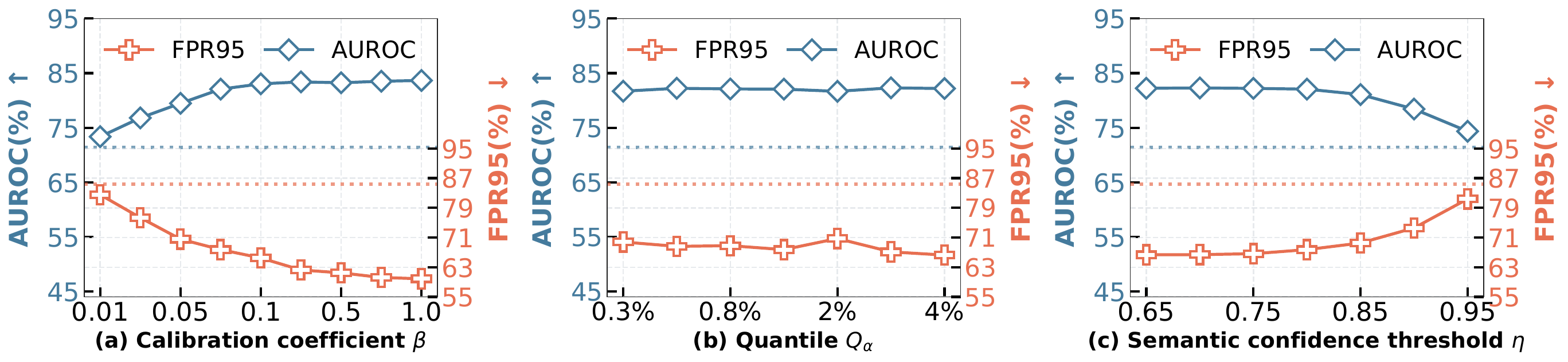}
        \end{center}
        \vspace{-13pt}
        \caption{
        Hyper-parameter sensitivity studies on ISIC-4.
Dashed lines represent the performance of the best baseline (i.e., OODD). }
        \label{fig:sensitivity}
        
\end{figure*}

\noindent \textbf{Sensitivity Study.}
We analyze the robustness of the proposed framework with respect to key hyperparameters.
Specifically, the calibration coefficient $\beta$ is varied within $[0.01, 1.0]$, the quantile $Q_{\alpha}$ within $[0.3\%, 4\%]$, and the semantic confidence threshold $\eta$ within $[0.65, 0.95]$.
As shown in Fig.~\ref{fig:sensitivity}, the proposed method consistently outperforms the best baseline across broad hyperparameter ranges, indicating stable performance and low sensitivity to hyperparameter choices.

\section{Conclusion}
We introduced CHILD, a training-free human-in-the-loop framework for OOD detection 
motivated by clinical deployment.
Under a strict feedback budget, CHILD selectively queries high-risk samples and incorporates sparse physician feedback through lightweight score-level calibration.
Extensive experiments on multiple medical benchmarks demonstrate that CHILD consistently improves OOD detection reliability under limited feedback.
These results support CHILD as a practical step toward safer medical AI systems with human oversight.

\begin{sloppypar}
\noindent\textbf{Acknowledgments.} This work is supported in part by the National Natural Science Foundation of China (grant No. 62571559), the Major Key Project of PCL (grant No. PCL2025AS209), and Guangdong Excellent Youth Team Program (grant No. 2023B1515040025).
\end{sloppypar}
\noindent\textbf{Disclosure of Interests.} Authors have no competing interests in the paper.

\bibliographystyle{splncs04}
\bibliography{Paper-1242}






\end{document}